\documentclass{article}
\usepackage{spconf,amsmath,graphicx}
\usepackage{xcolor}
\usepackage{caption}
\usepackage{subcaption}  
\usepackage{amssymb}% http://ctan.org/pkg/amssymb
\usepackage{pifont}% http://ctan.org/pkg/pifont
\newcommand{\cmark}{\ding{51}}%

\newcommand{\Rmnum}[1]{\expandafter\@slowromancap\romannumeral #1@}
\name{Yu-Jhe Li\hspace{1pt}\textsuperscript{*}\textsuperscript{1}, Hsin-Yu Chang\hspace{1pt}\textsuperscript{*}\textsuperscript{2}, 
Yu-Jing Lin\hspace{1pt}\textsuperscript{2}, Po-Wei Wu\hspace{1pt}\textsuperscript{2}, and Yu-Chiang Frank Wang\textsuperscript{1}\thanks{* Indicate equal contribution. }}
\address{
\textsuperscript{1}\hspace{1pt}Graduate Institute of Communication Engineering, National Taiwan University\\
\textsuperscript{2}\hspace{1pt}Department of Computer Science \& Information Engineering, National Taiwan University
}
\begin{document}

% \title{Deep Learning for Malicious Flow Detection}
\title{Deep Reinforcement Learning for Playing 2.5D Fighting Games}

% \author{\IEEEauthorblockN{Yu-Jhe Li\IEEEauthorrefmark{1},
% Yu-Jhe Li\IEEEauthorrefmark{2}, XXX\IEEEauthorrefmark{2}, XXX\IEEEauthorrefmark{1}\IEEEauthorrefmark{2}\IEEEauthorrefmark{3}}
% \IEEEauthorblockA{\IEEEauthorrefmark{1} Department of Electrical Engineering, National Taiwan University}
% \IEEEauthorblockA{\IEEEauthorrefmark{2} Graduate Institute of Communication Engineering, National Taiwan University}}

%\ninept
%
\maketitle
\begin{abstract}
Deep reinforcement learning has shown its success in game playing. However, 2.5D fighting games would be a challenging task to handle due to ambiguity in visual appearances like height or depth of the characters. Moreover, actions in such games typically involve particular sequential action orders, which also makes the network design very difficult. Based on the network of Asynchronous Advantage Actor-Critic (A3C), we create an OpenAI-gym-like gaming environment with the game of Little Fighter 2 (LF2), and present a novel A3C+ network for learning RL agents. The introduced model includes a Recurrent Info network, which utilizes game-related info features with recurrent layers to observe combo skills for fighting. In the experiments, we consider LF2 in different settings, which successfully demonstrates the use of our proposed model for learning 2.5D fighting games.
% few researchers start working on 2.5D (2.5-dimension) fighting games in which game designers try to manipulate characters in 3D (3-dimension) mode, but the game scenes are displayed in 2D (2-dimension).

% Reinforcement learning has always been a popular research topic in artificial intelligence. It is applied to various fields, especially OpenAI game playing\cite{brockman2016openai} such as Pong and Breakout, most of which are known as Atari 2600 games. Besides, more complicated games like DotA2 and StarCraft, containing 3D observations and various information in a game, are in process of advanced researches. However, few works were done on the domain of 2.5 dimensional games and fighting games. In a 2.5D game, depths of characters are important but hard to be detected, or recognized, by a neural network. On the other hand, a fighting game requires accurate control to the character. We hence introduce a deep reinforcement learning with a {\bf combined recurrent and information network }, which indicates positions of characters in a 2.5D game for the original deep convolutional network.

\end{abstract}
\section{Introduction}
\label{sec:intro}

Deep reinforcement learning (RL) has been widely utilized in recent topics and applications in computer vision, robotics, and machine learning. One of the successful examples is the Atari 2600 games. More specifically, Schaul et al. \cite{schaul2015prioritized} and Nair et al. \cite{nair2015massively} advance Deep Q Network (DQN) for learning Atari games. Recently, Schulman et al.~\cite{schulman2017proximal} and Gu et al. \cite{gu2016q} advance policy gradient to improve the training efficacy. Some other works \cite{hausknecht2015deep,todorov2012mujoco,christiano2017deep,kulkarni2016hierarchical} have also made great success in fine-tuning models for deep reinforcement learning.

% \cite{kulkarni2016hierarchical} use hierarchical DQN to learn the subgoals, which is applied to the Atari games with delayed rewards. 

% \cite{hausknecht2015deep} memorize previous observations in some aspects by adding Long-Short-Term-Memory (LSTM).\cite{christiano2017deep} learned complex goals like backflip on {\bf Mujoco \cite{todorov2012mujoco} } by human preference, and \cite{ho2016generative} use discriminator network to learn the expert's actions and train policy network with it. However, we're working on a 2.5D fighting game that they've not done.

With impressive progress on the learning of Atari games, games in more complex forms are further explored by the researchers. For example, Wu and Tian \cite{wu2016training} apply different deep RL models to learn first-person shooting games, and Tessler et al. \cite{tessler2017deep} utilize transfer learning for lifelong learning problems in Minecraft. Also, Silver et at. \cite{silver2016mastering} use the deep neural network to learn GO and defeat the world-class players. Leibo et al. \cite{leibo2017multi} further investigate the behaviors of the learned agents, and apply DQN for train multi-agent games. Some non-2D works contain a trial of 3D first player shooting (FPS) game {\bf Doom} \cite{kempka2016vizdoom}. Lample et al. \cite{lample2017playing} designed their new model to successfully train the agent in 3D games. Besides, Vinyals et al. \cite{vinyals2017starcraft} also extended their work on playing the non-2D strategy game: {\bf StartCraft 2}, and their improved method in training 3D game is also impressive. 

\begin{figure}[t]
\centering
\includegraphics[scale=.25]{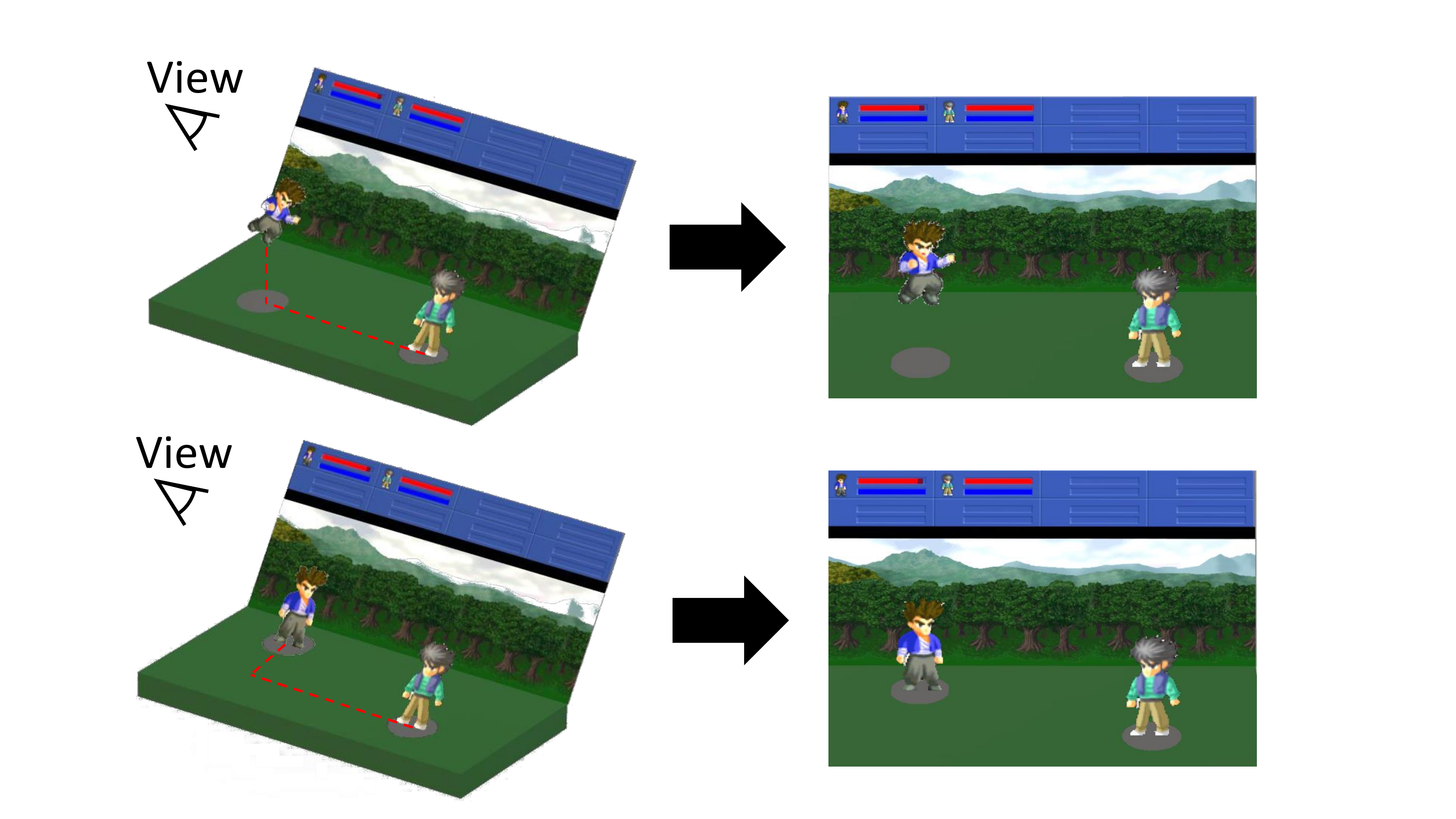}
\caption{
With orthographic projection, objects in varying distances in 2.5D games would exhibit the same size, which makes the learning of such games very challenging. Note that the left character in the upper image is jumping at a closer distance, while that in the lower one is at a farther distance. 
% Most of 2.5D game using orthographic projection in its viewpoint, meaning the size of image on the screen is irrelevant to the vertical distance between viewpoint and position of items. This figure shows that we can not distinguish the depth distance such as the difference between height (upper figure) and depth (lower figure) only from 2D scenes in 2.5D game.
} 
\end{figure}\label{fig1}

Nevertheless, existing deep RL models are typically designed for handling games in 2D or limitedly in 3D. Extension of such learning models for games beyond 2D would not be trivial. In this paper, we focus on learning 2.5D fighting games, in which characters are manipulated in 3D mode while they are displayed in 2D. We note that, in most 2.5D games, 3D objects in a scene are orthographically projected to the 2D screen. Thus, the size of the object in a farther distance would still be similar to that of a closer one. Fig.~\ref{fig1} shows example frames from a 2.5D fighting game: Little Fighter 2 (LF2) \cite{LF2}. We see that, the ambiguity between distance and height brings difficulties in training agents. Thus, utilizing the locations of the characters and memorizing their temporal states become vital for RL training. Moreover, characters in LF2 are with different combinations of skills, which make the learning even more challenging. As a result, direct uses of existing RL models would not be applicable.

In this paper, we present a novel RL architecture based on the Asynchronous Advantage Actor-Critic (A3C) network~\cite{mnih2016asynchronous}, and focus on learning the 2.5D fighting game of LF2. In addition to input screen frames, our proposed network takes a variety of information as input features. Together with the long short-term memory (LSTM) components, this allows us to exploit information across and beyond input screen frames, while the aforementioned ambiguity and challenges can be properly addressed. We refer to as proposed network as A3C+, and our main contributions are summarized as follows:

\vspace{-2mm}
\begin{itemize}
\setlength\itemsep{1pt}
\item We build an OpenAI-gym-like 2.5D game environment in which we successfully wrap the game server for agent learning and interaction.
\item Based on A3C, we propose an A3C+ which takes not only screen frames but also game-related features in a recurrent learning architecture.
\item We show that our proposed architecture favors 2.5D games, supported by quantitative and qualitative evaluation results.
% \item We completely built the OpenAI-gym-like 2.5D game environment, and then defined the action space and the reward method
% \item We proposed customized R-A3C (Recurrent Asynchronous Advantage Actor-Critic) and Info network which makes agent achieve high score in 2.5D fighting game
% \item We further specifically show that the Info network plays a essential role in overcoming the challenges in 2.5D fighting game
\end{itemize}
\vspace{-2mm}

\section{Environment Set-Up for 2.5D Fighting Game: Little Fighter 2 (LF2)}
\label{sec:format}

A major contribution of our work is the environment set-up for the freeware PC 2.5D fighting game of LF2~\cite{LF2}. As depicted in Fig.~\ref{fig2}, this process consists of three parts: game server, wrapper, and the agent. We apply the open-source Project F \cite{FLF}, which is a web-based implementation of LF2, as our game server. Next, we wrap this game into an OpenAI-gym-like environment \cite{brockman2016openai} \textit{env} with Python and web drivers (PhantomJS). The set-up details can be found at~\cite{lf2gym}. 
% \subsection{Environment construction}
\begin{figure}[t]
\centering
\includegraphics[scale=.5]{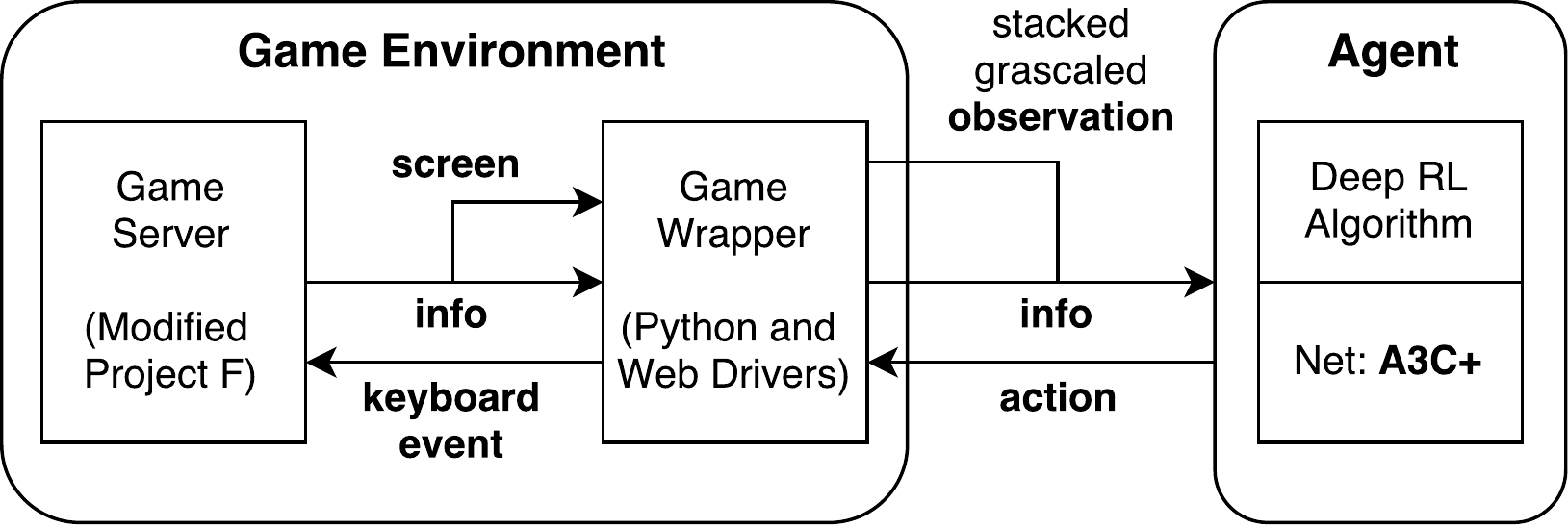}
\caption{Our framework describing the game environment and the agent. The agent interacts with the game server through the game wrapper, which makes allows the agents to take actions and obtain rewards from the environment.}
\end{figure}
\label{fig2}

\begin{figure*}[t]
    \centering
    \includegraphics[scale=.33]{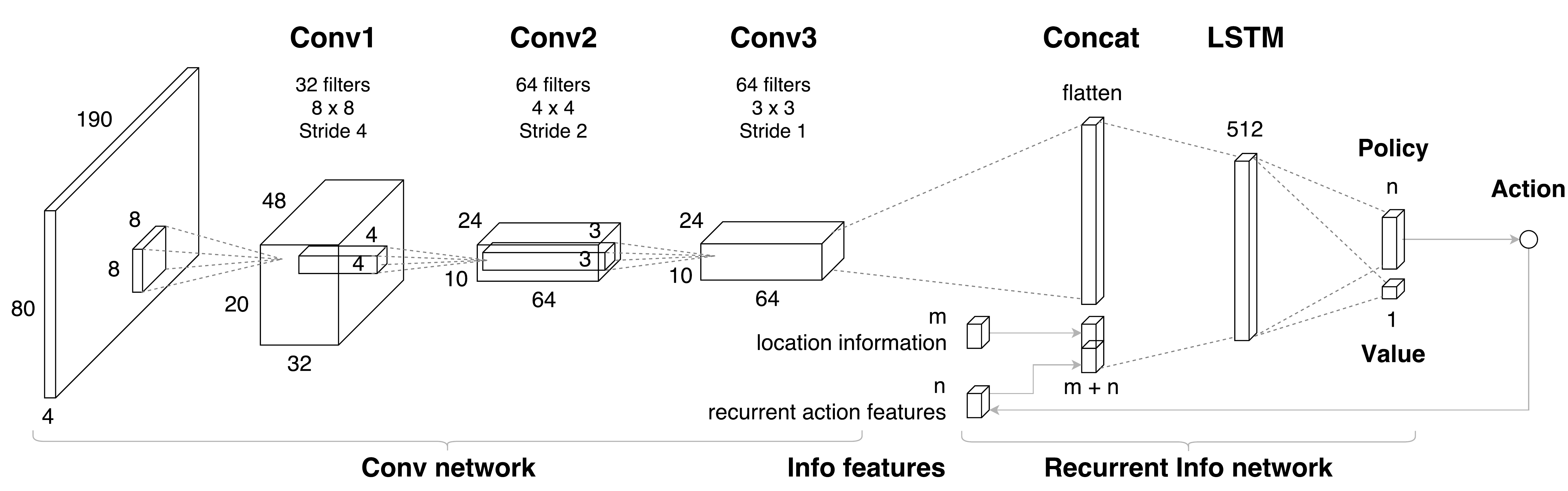}
    \caption{Our network of A3C+. Note that the introduced recurrent info network takes 2.5D game-related and fighting-action related features, aiming to better describe game environments and to allow our agent for winning the fighting game.}
    \label{fig3}
\end{figure*}

\subsection{Observations for RL}
In order to learn the proposed RL-based network architecture, an array of 4-stacked, gray-scaled successive screenshots is given as the observed state to the agent at each step, which reveals the motion information of the agents in a scene. In order to perform this process efficiently, we apply the frame-skipping technique of~\cite{bellemare2013arcade}, with skip frame parameter $k$ set and fixed as 4. To be more precise, the agent learned by our model would take an action at every four successive frames.% make sure

% To have efficient training process, when the agent takes one action or we resets the environment, we give an array of 4-stacked, gray-scaled screenshots as observed states to the agent. We also use the frame-skipping technique which Bellemare et al. \cite{bellemare2013arcade} proposed and set the skip frame parameter $k$ to 4.

\subsection{Action Space and Rewards}

% Due to the 4-skip technique, each action is also defined as skip-4 action, which is a sequence of four basic actions, or keyboard events. %???

The action space of LF2 comprises 8 basic actions and 8 combo skills. While \textit{idle, up, right, down, left, attack, jump, }and \textit{defend} constitute the basic action, combo skills are represented by sequences of basic actions. The role \textit {Firen}, for example, makes explosion by taking \textit{defend, up,} and \textit{jump} successively. We note that, a set combo skills with 8 different combinations are defined as [\textit{defend} + \textit{up/right/down/left} + \textit{attack/jump}], while different kinds of combo skills are available for different characters.

In 2.5D fighting games, the goal is to defeat the adversary character in terms of the remaining health points (HP) of the two characters. We define the reward as such differences. Thus, a positive reward is given by any damage to the opponent character, while any injury to our agent results in a negative reward. Thus, our agent would learn through the training process by taking proper actions to beat the opponent.
% In a 2.5D fighting game, although it is composed of a bunch of complicated action controls an agent can take in each step, the final goal is to defeat the adversary. Therefore, we take the difference of health points (hp) between two states as our reward. Any damage to the opponent gives rise to a positive reward for the agent while any injury to the agent results in a negative reward.

% \subsection{
% The main challenge of this game is that we need to learn the position information under the complex environment and present which action results in positive rewards. We train our agents against built-in rule-based AI, which is smart and aggressive at the difficulty of \textit{CHALLANGAR} \cite{FLF}. % 這邊說smart and aggressive讓別人覺得對手很強xDD

% Need more modifications to make it smoothly......
\vspace{-2mm}
\section{Our Proposed Model}

\subsection{Brief Review of Asynchronous Advantage Actor-Critic (A3C)}
% [DQN part begin]
% \subsubsection{Deep Q Learning}
% Deep Q-Learning \cite{watkins1992q}\cite{mnih2015human} estimates the value of executing an action from a given state. Such value estimates are referred to as state action values, or sometimes simply Q-values. The DQN(Deep Q-Learning network) model, which is a neural network parameterized by weights and biases collectively denoted as $\theta$, is used to approximate the Q-values. Updates are made to the parameters of the network to minimize a differentiable loss function:

% \begin{equation} \label{eu_2}
% L(s,a|\theta_{i}) = (r+\gamma \max_{a'}Q(s',a'|\theta_{i})-Q(s,a|\theta_{i}))^2
% \end{equation}
% [DQN part end]

% Instead a model is used to approximate
% the Q-values (Mnih et al. 2015). In the case of Deep Q Learning,
% the model is a neural network parameterized by
% weights and biases collectively denoted as $\theta$. Q-values are
% estimated online by querying the output nodes of the network
% after performing a forward pass given a state input.
% Such Q-values are denoted $Q(s, a|\theta)$. Instead of updating individual
% Q-values, updates are now made to the parameters
% of the network to minimize a differentiable loss function:

% \subsubsection{Learning algorithm}

Since our proposed deep RL architecture is based on A3C~\cite{mnih2016asynchronous}, we now briefly review this state-of-the-art RL model for the sake of completeness. For deep RL, the agent is learned via a deep neural network with parameter $\theta$, which defines a policy $\pi_{\theta}$ and an estimate of the value function $V_{\theta}$. The agent receives observation $s_t$ at each time step $t$, selects an action $a_t$ with probability $\pi_\theta(a_t|s_t)$, and then receives a reward $r_t$ from the game environment. The goal of the RL agent is to maximize the return $R_t=\sum_{k=0}^{\infty}\gamma^kr_{t+k+1}$, where $\gamma$ is a discount factor. The parameters of the policy and the value are learned using Asynchronous Advantage Actor-Critic (A3C) described by Mnih et al. \cite{mnih2016asynchronous}. A3C is a policy gradient method with an approximate gradient ascent on $\mathbf{E}[R_t]$. The update can be seen as $\bigtriangledown_{\theta^{'}}log\pi_{\theta}(a_t|s_t;\theta)A(a_t,s_t;\theta)$, in which $A(a_t,s_t;\theta)$ is an estimate of advantage function given by $R_t-V(s_t;\theta)$, where $V(s_t;\theta)$ is a value function estimate of the expected return $\mathbf{E}[R_t]$. Based on the recent models of~\cite{mnih2016asynchronous,williams1990efficient}, an additional entropy regularization is applied, with the complete version of the objective function as $\bigtriangledown_{\theta^{'}}log\pi_{\theta}(a_t|s_t;\theta)(R_t-V(s_t;\theta))+\beta\bigtriangledown_{\theta^{'}}\pi_{\theta}(s_t;\theta^{'})log\pi_{\theta}(s_t;\theta^{'})$. We note that, the hyperparameter $\beta$ controls the strength of the entropy regularization term. 

% defined gradient is detailed as:
% \[
% \underbrace{(R_t-v_\theta(s_t))\bigtriangledown_{\theta}log\pi_{\theta}(a_t|s_t)}_{policy gradient}+\underbrace{\beta(R_t-v_\theta(s_t))\bigtriangledown_{\theta}v_\theta(s_t)}_{value gradient}
% \label{eu_3}\]

% \begin{equation} \label{eu_3}
% \underbrace{(R_t-v_\theta(s_t))\bigtriangledown_{\theta}log\pi_{\theta}(a_t|s_t)}_{policy-gradient}+\underbrace{(R_t-v_\theta(s_t))\bigtriangledown_{\theta}v_\theta(s_t)}_{value-gradient}
% \end{equation}

% \subsubsection{Recurrent layer}

% A recurrent layer is found effective in reinforcement learning on games.

% 將這段搬動到3.2.2以敘述recurrent layer
% Moreover, Mnih et al. \cite{mnih2016asynchronous} compared naive A3C and its recurrent version, and found that the latter exhibits improved performance and thus is preferable. In addition, Hausknecht and Stone \cite{hausknecht2015deep} also introduce a deep recurrent Q network, which is a improved DQN network with recurrent layers. In our proposed A3C+ network (depicted in Fig.~\ref{fig3} and detailed later), we found it would be preferable memorize previous states, so we also advance the recurrent module and will confirm its effectiveness later.

% Besides, in the Little Fighter 2, a 2.5D fighting game, players take complicated actions to control characters appropriately. Combo skills are also supposed to be learned in the training process. We consider it helpful that a recurrent layer memorizes previous states and actions the agent took. Hence, the LSTM replaces the original fully-connected layer in the Info network.

\begin{figure}[t]
	\centering
	\begin{subfigure}{0.4\textwidth} % width of left subfigure
    	\centering
		\includegraphics[width=\textwidth]{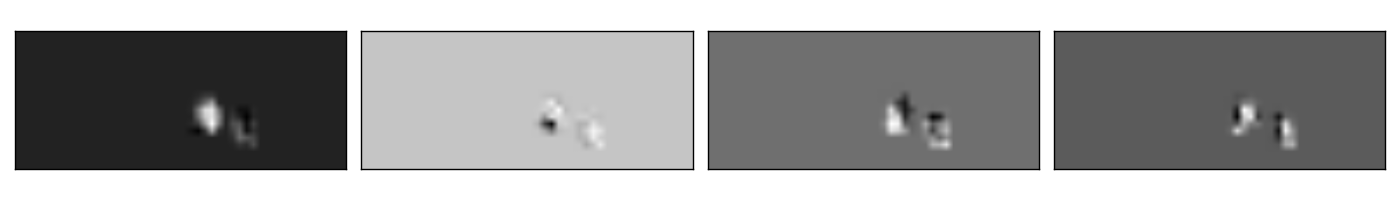}
% 		\caption{Conv1 feature maps} % subcaption
	\end{subfigure}
	
	\begin{subfigure}{0.4\textwidth} % width of right subfigure
    	\centering
		\includegraphics[width=\textwidth]{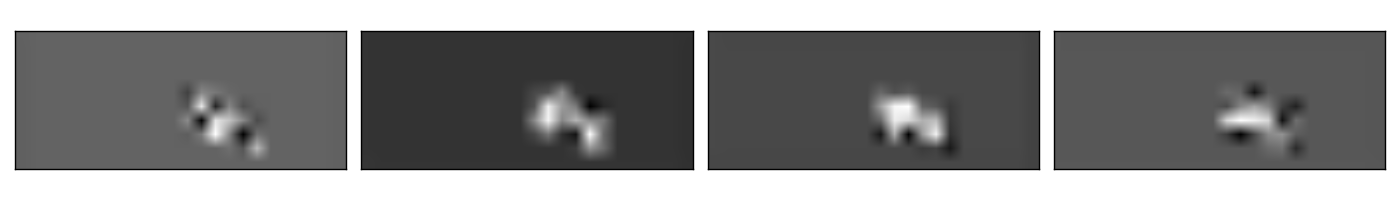}
% 		\caption{Conv2 feature maps} % subcaption
	\end{subfigure}
    
	\begin{subfigure}{0.4\textwidth} % width of right subfigure
    	\centering
		\includegraphics[width=\textwidth]{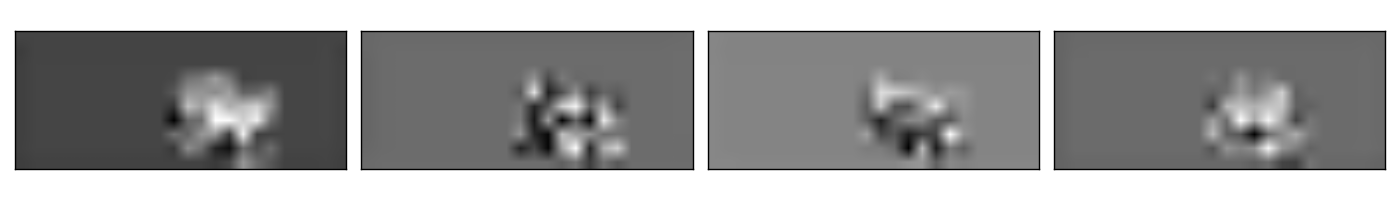}
% 		\caption{Conv3 feature maps} % subcaption
	\end{subfigure}
	\caption{Example visualization of convolution layers in Conv network for LF2 using standard A3C (top to bottom: Conv1, Conv2, Conv3 respectively). } 
\label{fig4}
\end{figure}

% A3C Experiments Figure
\begin{figure*}[t]
	\centering
	\begin{subfigure}{0.49\textwidth} % width of left subfigure
		\includegraphics[width=\textwidth]{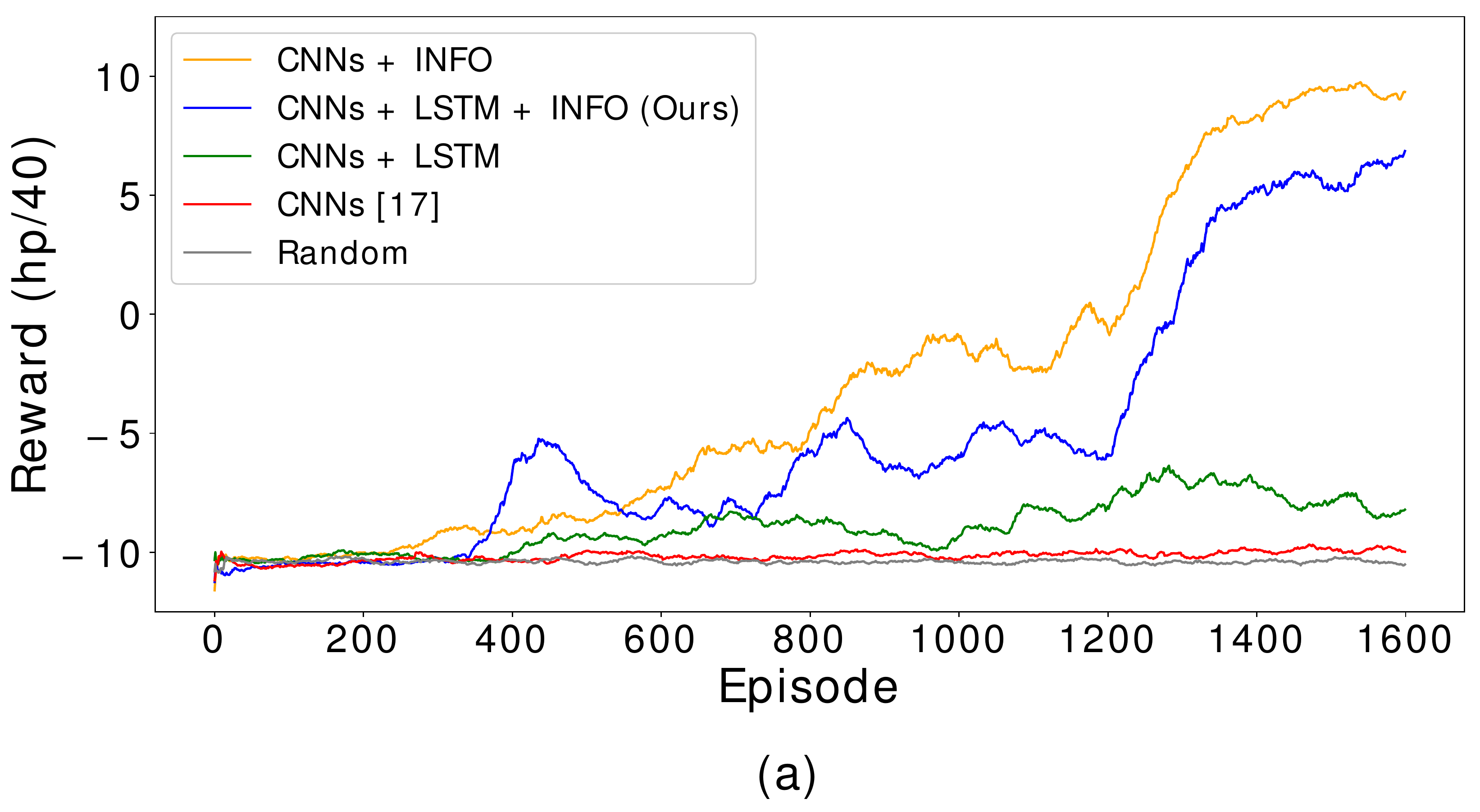} 
%         \caption{Basic Setting}
	\end{subfigure} 
	\begin{subfigure}{0.49\textwidth} % width of right subfigure 
		\includegraphics[width=\textwidth]{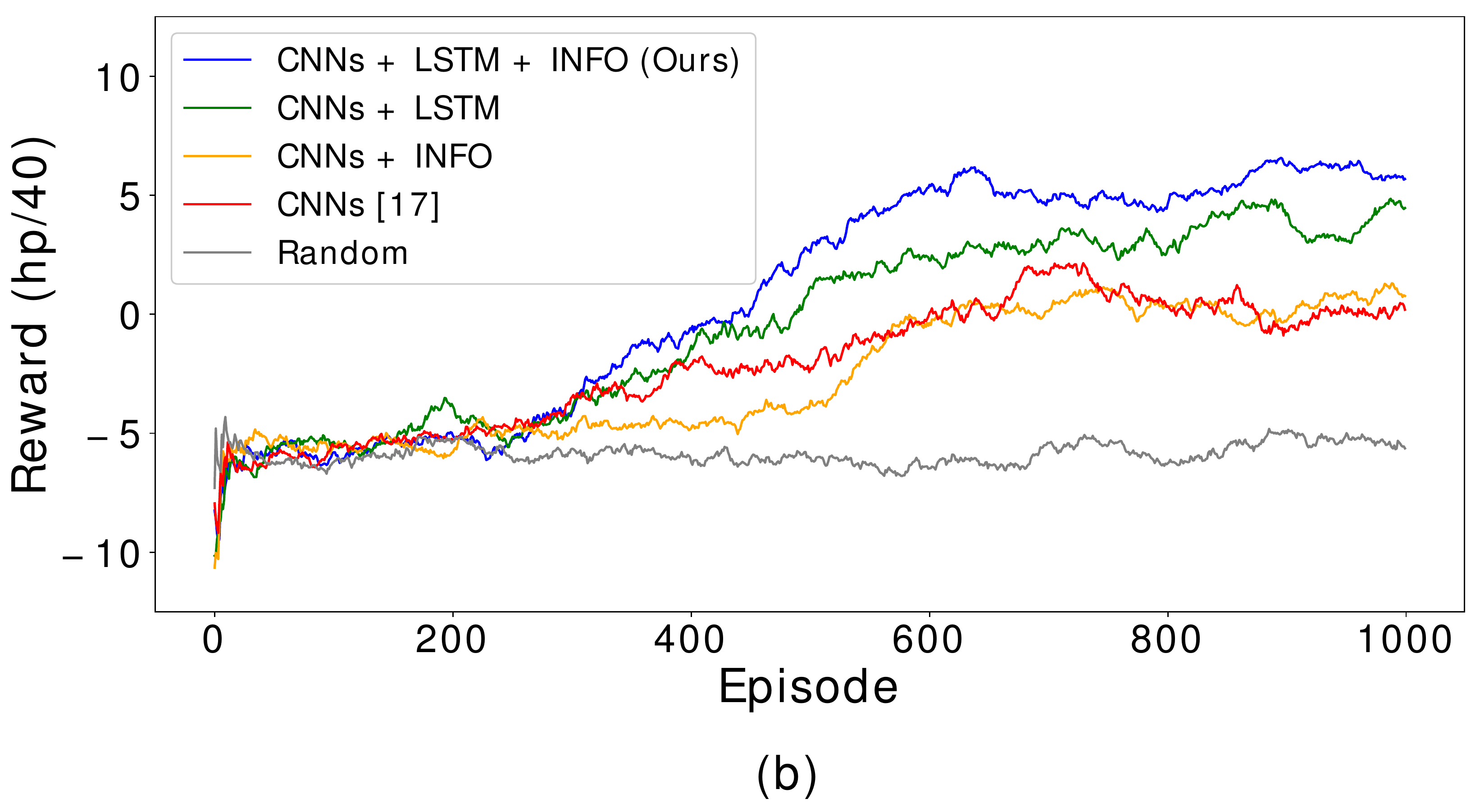} 
%         \caption{Advanced Setting}
	\end{subfigure} 
%     \begin{subfigure}{0.3\textwidth} % width of right subfigure 
% 		\includegraphics[width=\textwidth]{DQN_davis.eps}
% % 		\caption{Advanced Setting} % subcaption
% 	\end{subfigure}
  \vspace{-1mm}
	\caption{Average rewards over episodes when training A3C models in (a) basic and (b) advanced settings. } % caption for whole figure
    \vspace{+4mm}
    \label{fig5}
\end{figure*}

% The basic convolution neural network has been widely used when training agent to play game in Atari game environment \cite{brockman2016openai}. 

% Since its well-designed architecture, related works have shown its success in the domain of game playing.
\subsection{Our A3C+ for Learning 2.5D Fighting Games}
With A3C applied to basic 2D games like Atari, how to reflect features vital for 2.5D games (e.g., character location) is still a challenging task. In our proposed network of {\bf A3C+}, which extends A3C and includes a novel {\bf Recurrent Info network} (i.e., recurrent layers with introduced 2.5D game-related features) as illustrated in Fig.~\ref{fig3}. We now discuss the A3C+ components in the following subsections.

% The Conv network is referred to the widely used Atari network, which is successfully applied to various games, except for 2.5D fighting games.
% The Recurrent Info network takes the feature map from the former network as well as game-related features, and generates policy and value as its prediction. More detailed ideas of design will be discuss in the following We note that, parts.

\subsubsection{LSTM for recurrent learning}
Previously, Hausknecht and Stone \cite{hausknecht2015deep} presented a deep recurrent Q network, which improves DQN with recurrent layers for observing and learning temporal information. Similar ideas have been utilized in standard A3C~\cite{mnih2016asynchronous}, showing preferable results in RL tasks. In view of the above success, we also advance recurrent layers in our A3C+. More specifically, we utilize the LSTM with the observation of the features related to 2.5D games (see Sect. 3.2.2).

% Although the basic convolution neural network has long been successful in Atari game environment, the feature maps obtained from the basic network can't reflect the exact position information since the agent can't discriminate between itself and the opponent. From Fig. 4, neither one of these three convolution layers can exactly tell who is the rule-based AI.

\subsubsection{Info features for 2.5D games}
% \subsubsection{Position feature}
As noted earlier, standard A3C cannot easily describe game-related features like identification, actions, or height/distance of characters (see Fig.~\ref{fig4} as example). To solve these problems, we take the {\bf location information} of both our agent and the opponent into consideration. To be more precise, we extract the coordinates of both characters from the game environment, as well as their heights, depths, the distance between them. These representations are concatenated with the CNN visual feature, which would be the inputs to the LSTM module as depicted in Fig.~\ref{fig3}. We believe that, when learning complicated 2.5D fighting games, it is vital to allow the design network to observe the above game-related features.

\subsubsection{Actions of the agents in info features}
In fighting games like LF2, effective hits on the opponents involve sequential orders actions (e.g., an agent needs to jump to the back of the opponent and then punch the opponent immediately to gain the highest reward). To better train our agent using the proposed network, we particularly consider {\bf recurrent action features} as parts of the inputs to the recurrent info network. With such joint feature observation and the use of LSTMs, we aim at learning proper 2.5D game-related and action-order related features, which would allow our agent to survive longer and/or defeat the opponent.

\section{Experiments}
\label{sec:majhead}

\subsection{Implementation Details and Settings}
% The model structure can be found in the previous section and in Fig. 3. For more details about out experimental setting, we used leaky ReLU \cite{maas2013rectifier} as our activation function throughout the whole network. 
% "activation function" For more details about out experimental setting, we used leaky ReLU \cite{maas2013rectifier} as our activated function throughout the whole network. 

% We utilized RMSprop \cite{tieleman2012lecture} as optimizer with decay $0.99$, learning rate $0.0001$, and the discount factor is $\gamma = 0.99$ in the policy learning process. 

%These setting are simple, we didn't spend lots of time on adjusting them that we could get pretty good results. 
As mentioned earlier, we consider Little Fighter 2 (LF2) as the 2.5D fighting game in our work. In LF2, each character has different skills in terms of offense. For evaluation simplicity, we consider the role of \textit{Davis}, which is the main character of this game, as our training agent. To confirm the effectiveness of the proposed {\bf Recurrent Info network} in our A3C+ network, we take the {\bf Conv network}~\cite{mnih2016asynchronous} in the standard A3C as the baseline network. And we perform ablation tests to verify the contributions of the proposed network components (i.e., uses of info features and recurrent layers of LSTM).
% To verify that the {\bf Info network} work in this task, we use the {\bf Conv network} as baseline network. Concatenate Info features after Conv network and compare it with which without Info features. 
%After concatenating, 

In our evaluation, two different settings are considered:
\vspace{-1mm}
\begin{itemize}
\setlength\itemsep{1pt}
\item Basic: Only 8 basic actions allowed (see Sect. 2.2). 
\item Advanced: 8 actions in the basic setting, plus 8 combo skills without constraints on the magic points (MP). 
\end{itemize}
\vspace{-1mm}
\noindent We note that, for the advanced setting, each LF2 agent has a total of 8 combo skills. When utilizing our A3C+ for such fighting games, we wish to verify that the agent would not only learn the basic actions, it would learn the correlation between them and produce combo skills as well. As for the opponent in the game, we consider a rule-based AI. That is, it is always able to use all 16 actions in either of the two settings.

% However, most characters have fewer than eight valid combo skills, which means, some combo skill actions bring no effect for those characters. 

% By the way, most characters have less than eight valid combo skills, which means some combinations of actions brings no effect for those characters. 
% , where the agent is "taught" to use combo skill rather than learning correct sequence of that from scratch. 

% Note that the opponent in the game, which is the rule-base AI, is always able to use total 16 actions in each of the setting.

% \begin{table}[t] 
% \centering
% \caption{Winning Rate (WR) on testing the final model of different train-set.}
% \label{my-label}
% \begin{tabular}{l|cc|cc}
%  & \multicolumn{2}{c|}{Model Setting} & \multicolumn{2}{c}{Game Setting} \\
%  & INFO & LSTM & Basic A3C & Advanced A3C \\ \hline
% Train-set 1 & \cmark & \cmark & {\bf 99.5\%}  & {\bf 95.9\%}  \\ \hline
% Train-set 2 &  & \cmark & 3.9\% & 86.3\% \\ \hline
% Train-set 3 & \cmark &  & 99.3\% & 61.3\% \\ \hline
% Train-set 4 &  &  & 0.0\% & 57.6\% \\ \hline
% Random &  &  & 0.0\% & 10.6\%
% \end{tabular}
% \end{table}

\subsection{Evaluation}
We start with health points (HP) as 500 for both our agent and the opponent, with reward scaled to $1/40$ of the HP differences. In other words, the highest reward our agent can achieve per episode is between -12.5 and +12.5. To make the evaluation comparisons more complete, we also consider a baby-agent, which simply takes a random action at each step and thus can be viewed as a naive baseline model.
% \vspace{-4mm}
\subsubsection{Training stage}
Fig.~\ref{fig5} shows the rewards observed by different models using basic and advanced settings. In Fig.~\ref{fig5} (a), our agent was only trained to take 8 basic actions, and no combination skills were allowed. Thus, the use of info features in our A3C+ was sufficient to achieve promising rewards, when comparing to the use of our full model (i.e., info features plus LSTM recurrent layers). More importantly, advancing recurrent layers or not, models without observing info features would not be able to produce satisfactory results. This confirms that, the use of 2.5D game-related features would be beneficial in learning 2.5D fighting games, and our proposed network architecture with such features would be preferable.

% The reason is probably because recurrent action features provide enough information for our agent in this setting, such that the model with the recurrent layer (LSTM) or not does not have a significant impact to the result. 

% Fig. 5 (a) shows that the obtained rewards are always less than 0 in the beginning, which indicates that agents always lose in the early period. However, after 1200 episodes, agents training with the Info vector both learn to defeat the rule-based AI, showing that the game-related features take an important role in playing 2.5D fighting game, also indicating the positive effectiveness of position features and recurrent action features. On the other hand, agents without our Recurrent Info network perform much worser. 
% %We conclude that the model with INFO network grants the agents more helpful and extra information than the basic network with only the feature maps. 
% Furthermore, in this setting the model without the Info vector but with the LSTM cell still can not learn well to beat opponent. The reason is probably because the recurrent layer (LSTM) provides information of previous actions, which is the most important feature in our basic setting of fighting game. 

On the other hand, Fig.~\ref{fig5} (b) shows the agent training in the advanced setting. From this figure, we observe that the use of both info feature and the LSTM recurrent layers would be preferable, since this full version of our A3C produced the highest rewards compared to other baseline models or controlled experiments. This is because that, the combo skills may receive rewards after several screen frames, and thus delayed rewards are more likely to be observed and taken into consideration for the actions in the past. The network models with only info features but without the recurrent layer can only observe the information from the previous action, which would not be sufficient produce preferable actions or combo skills. In other words, when combination skills are allowed, it would be preferable to learn actions in proper sequential orders so that it becomes more likely to defeat the opponent. 
% kind of complicated fighting game
% The two agents without LSTM have worse performance than agents with LSTM, which is because the rewards we get will delay after the agent perform skills. To train the delayed rewards game environment, we need to use LSTM network to trace which action we perform can get rewards after several frames. 

\begin{table}[t] 
\centering
\caption{Winning rate (WR) of the deep RL agents trained in different settings.}
\label{my-label}
\begin{tabular}{l|cc|cc}
 & \multicolumn{2}{c|}{Model Setting} & \multicolumn{2}{c}{Game Setting} \\
Agent & LSTM & INFO & Basic & Advanced \\ \hline
Ours & \cmark & \cmark & 95.5\%  & {\bf 95.9\%}  \\ %\hline
Ours w/o LSTM & & \cmark & {\bf99.3\%} & 61.3\% \\ %\hline
CNNs w/ LSTM & \cmark & & 11.9\% & 86.3\% \\ %\hline
CNNs ~\cite{mnih2016asynchronous} &  &  & 7.3\% & 57.6\% \\ %\hline
Random &  &  & 4.9\% & 10.6\%
\end{tabular}
\label{tab1}
\end{table}
\vspace{-0mm}

\subsubsection{Testing stage}
In addition to the comparisons of the observed rewards during training, we further present the evaluation of testing. To be more specific, we present the winning rate (WR) of the learned agent, based on trained models for 1000 episodes. The results are listed in Table~\ref{tab1}. Recall that the opponent in our experiment is a rule-based AI taking both basic actions and combo skills during fighting. From Table~\ref{tab1}, we see that the use of the full version of our A3C+ achieved the most satisfactory WR. For the basic setting (i.e., only basic actions allows without combo skills), the use of recurrent layers did not exploit additional sequential action information, and thus the use of info features would be sufficiently satisfactory (i.e., WR as 99.3\%).
From the above experiments, the effectiveness of our A3C+ for learning 2.5D fighting games can be successfully verified.

% is sufficient to achieve promising rewards, when comparing to the
%Thus, it is expected that the agent trained by the full version of our A3C+ (i.e., info features plus LSTM as recurrent info network) produced the highest WR for both basic and advanced settings. 

% Therefore, we use the final model that training above to test for 1000 episodes, results are shown in Table 1. We can see the winning rate of agent with INFO and LSTM can be over 95\% on each of the setting, and the agent with INFO network always perform better than the one without.

% [Original]
% Subheadings should appear in lower case (initial word capitalized) in
% boldface.  They should start at the left margin on a separate line.
% \subsubsection{Sub-subheadings}
% \label{sssec:subsubhead}

% [Original]
% Sub-subheadings, as in this paragraph, are discouraged. However, if you
% must use them, they should appear in lower case (initial word
% capitalized) and start at the left margin on a separate line, with paragraph
% text beginning on the following line.  They should be in italics.
\vspace{0mm}
\section{Conclusion}
As the first to advance deep reinforcement learning for 2.5D fighting games, our proposed network extends Asynchronous Advantage Actor-Critic (A3C) and exploits game-related info features with recurrent layers. This not only alleviates the problem of visual ambiguity of the characters during fighting, it also allows us to observe proper action orders for producing combo action skills. With our set-up of game environment and the game of LF2, our experiments confirmed that the use of our game-related info features is crucial in 2.5D fighting games, while the full version of ours (i.e., with the proposed recurrent info networks) produces the best winning rate and performs favorably against recent or baseline deep reinforcement learning models.

% Apart from the related improvements considering game features on RL, which makes agent understand situations in 3D games, our result justifies that the proposed method is able to locate characters and overcome the difficulty of depth ambiguity in the 2.5D environment. 
% Experimental result demonstrates that our improved network do help the agent achieve high score in either basic or advanced setting, and it justifies that the proposed method is able to overcome the difficulty of depth ambiguity. 

% justified that the Info network plays an essential role in overcoming the challenges such as depth ambiguity in 2.5D fighting games.
% In this paper, we build a gym-liked 2.5D fighting game environment which provides a friendly application programming interface for training. 
% Our proposed integration of basic A3C with INFO network make our agent learn state more precisely from extra features. 
% We also validated the effectiveness of LSTM in 2.5D fighting game learning. This architecture leads to be capable of making better decisions and taking actions more smoothly on 2.5D fighting games. 
% 忘了留改之前的...
%Eventually, we successfully train on the 2.5D fighting game and get perfect result on our experiment.
\label{sec:print}

\newpage

\bibliographystyle{IEEEbib}
\bibliography{refs}

\end{document}